\documentclass[10pt,twocolumn,letterpaper]{article}

\usepackage{wacv}
\usepackage{times}
\usepackage{epsfig}
\usepackage{graphicx}
\usepackage{amsmath}
\usepackage{amssymb}
\usepackage{algorithm}
\usepackage{algpseudocode}
\usepackage{booktabs}
\usepackage{multirow}
\usepackage[labelfont=bf]{caption}

\usepackage{forloop}
\newcounter{ct}

\newcommand{\doubleQuote}[1]{\lq\lq{#1}\rq\rq}

\graphicspath{{figures/}}

\wacvfinalcopy % *** Uncomment this line for the final submission

\def\wacvPaperID{36} % *** Enter the wacv Paper ID here

\ifwacvfinal\fi
\begin{document}

%%%%%%%%% TITLE
\title{Learning Sports Camera Selection from Internet Videos}

% Authors at the same institution
%\author{First Author \hspace{2cm} Second Author \\
% Authors at different institutions
\author{Jianhui Chen $^*$  \qquad Keyu Lu $^\dagger$ \qquad Sijia Tian $^*$ \qquad James J. Little $^*$\\
$^*$University of British Columbia  \qquad $^\dagger$National University of Defense Technology\\
{\tt\small \{jhchen14, candice, little\}@cs.ubc.ca \qquad \tt\small keyu.lu@nudt.edu.cn }
%\and
%Second Author \\
}

\maketitle
\ifwacvfinal\thispagestyle{empty}\fi

%%%%%%%%% ABSTRACT
\begin{abstract}
This work addresses camera selection, the task of predicting which camera should be \doubleQuote{on air} from multiple candidate cameras for soccer broadcast. The task is challenging because of the scarcity of learning data with all candidate views. Meanwhile, broadcast videos are freely available on the Internet (\eg Youtube). However, these videos only record the selected camera views, omitting the other candidate views. To overcome this problem, we first introduce a random survival forest (RSF) method to impute the incomplete data effectively. Then, we propose a spatial-appearance heatmap to describe foreground objects (\eg players and balls) in an image. To evaluate the performance of our system, we collect the largest-ever dataset for soccer broadcasting camera selection. It has one main game which has all candidate views and twelve auxiliary games which only have the broadcast view. Our method significantly outperforms state-of-the-art methods on this challenging dataset. Further analysis suggests that the improvement in performance is indeed from the extra information from auxiliary games. 
\end{abstract}

%%%%%%%%% BODY TEXT
\section{Introduction}
The sports market in North America was worth 69.3 billion in 2017 and is expected to reach 78.5 billion by 2021. Our work focuses on soccer game which has about 40\% shares of the global sports market. As the biggest reason for market growth, media rights (game broadcasts and other sports media content) are projected to increase from 19.1 billion in 2017 to 22.7 billion in 2021 \cite{pwc2017at}.  Computational broadcasting is a promising way to offer consumers with various live game experiences and to decrease the cost of media production. Automatic camera selection is one of the key techniques in computational broadcasting. 

\begin{figure}[t]
	\centering
	\includegraphics[width=0.9\linewidth]{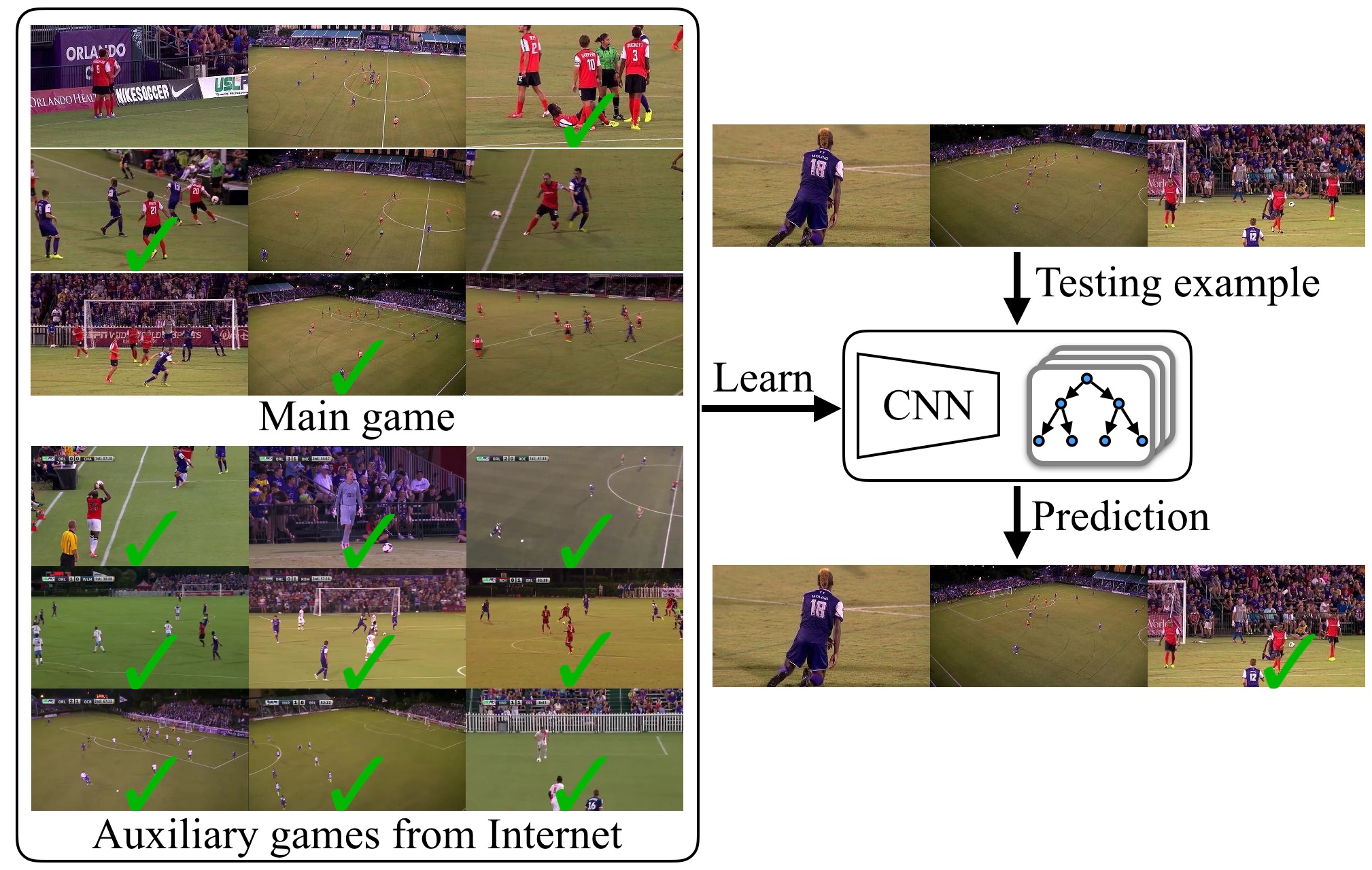}    
	\caption{Learning camera selection from Internet videos. The goal of our work is to select one camera from multiple candidate cameras for sports broadcast. The ideal way is trained from a dataset that has all candidate views such as the main game shown in the figure. However, it is hard to acquire this kind of data (including candidate videos and broadcasting videos) because they are generally not available to researchers (owned by broadcasting companies). Our method uses publicly available  Internet videos as auxiliary data to train a model with state-of-the-art prediction accuracy. Best viewed in color.}
    \vspace{-0.15in}
	\label{fig:main_idea}
\end{figure}

Machine learning has produced impressive results on view point selection. These include automatic \cite{chen2016learning,su2016activity,su2017making,hu2017deep360,chen2018camera} and semi-automatic \cite{foote2013one} methods from various inputs such as first-person cameras, static and pan-tilt-zoom (PTZ) cameras. The underlying assumption of these methods is that large training data are easily available. 

\textbf{Motivation} For sports camera selection, amounts of large training data are not directly available. As a result, most previous methods are trained on a single game (main game) because researchers can not acquire the data that are owned by broadcasting companies \cite{chen2013computational,chen2018camera}. On the other hand, broadcast videos are widely available on the Internet (\eg Youtube). These games (auxiliary games) provide a large number of positive examples. Using these Internet videos can scale up the training data with negligible cost.
 
In practice, arbitrarily choosing auxiliary games does not necessarily improve the performance, when main games are from minor leagues while auxiliary games are from premier leagues. So, the main game and the auxiliary games should be similar in terms of camera locations and the action of players. Although a universal camera selection model should be the final goal, a model for a specific team is also valuable. For example, teams in minor leagues can reduce the cost of live broadcasting for host games. Targeting these applications, we constrain the main games and auxiliary games to be from the same stadium at the current stage. 

The main challenge of using auxiliary games is the missing views in the video composition. Omitting non-broadcast views is the default setting for TV channels and live streams on the Internet. As a result, the amount of complete and incomplete data is highly unbalanced. To overcome this challenge, we introduce the random survival forest method (RSF) \cite{ishwaran2008random} from statistical learning to impute the missing data. To the best of our knowledge, we are the first to use Internet videos and RSF to solve camera selection problems.

The second challenge is from the potentially negative impact of background information in auxiliary games. Auxiliary games are very different in lighting, fan celebration and stadium decoration. In practice, camera operators are trained to capture interesting players and keep the ball visible \cite{owens2015television}. Inspired by this observation, we propose a spatial-appearance heatmap to represent foreground objects locations and their appearances jointly.

Our main contributions are: (1) Using Internet data and random survival forests to address the data scarcity problem in camera selection for soccer games. (2) Proposing a spatial-appearance heatmap to effectively represent foreground objects. With these novel techniques, our method significantly outperforms state-of-the-art methods on a challenging soccer dataset. While we present results on soccer games, the technique developed in this work can apply to many other team sports such as basketball and hockey.

\section{Related Work}
\textbf{Data Scarcity and Imputation} The availability of a large quantity of labeled training data is critical for successful learning methods. This assumption is unrealistic for many tasks. As a result, many recent works have explored alternative training schemes, such as unsupervised learning \cite{zhou2017unsupervised}, and tasks where ground truth is very easy to acquire \cite{agrawal2016learning}. We follow this line of work with additional attention to data imputation approaches.

Data imputation fills in missing data from existing data \cite{enders2010applied}. The missing data falls into three categories: missing at random (MAR), missing completely at random (MCAR) and missing not at random (MNAR). The missing data in our problem is MNAR because the missing data is related to the value itself (all missing data is unselected by human operators). Our solution is adapted from the state-of-the-art random survival forests method \cite{ishwaran2008random,tang2017random}.

\textbf{Camera Viewpoint Prediction} In single camera systems, previous techniques have been proposed to predict camera angles for PTZ cameras~\cite{chen2016learning} and to generate natural-looking normal field-of-view (NFOV) video from $360^o$ panoramic views \cite{su2016activity,su2017making,hu2017deep360}. In multi-camera systems, camera viewpoint prediction methods select a subset of all available cameras~ \cite{wang2008automatic,daniyal2011multi,tessens2014camera,arev2014automatic,fujisawa2015automatic,gaddam2014your,yus2015multicamba,wang2016personal,lefevre2018automatic}. In broadcast systems, semi-automatic \cite{foote2013one} and fully automatic systems have been developed in practice. For example, Chen \etal \cite{chen2013computational} proposed the automated director assistance (ADA) method to recommend the best view from a large number of cameras using hand-crafted features for field hockey games. Chen \etal \cite{chen2018camera} modeled camera selection as a regression problem constrained by temporal smoothness. They proposed a cumulative distribution function (CDF) regularization to prevent too short or too long camera durations. However, their method requires a real-valued label (visual importance) for each candidate frame. Our problem belongs to multiple dynamic camera systems. 

\textbf{Video Analysis and Feature Representation} Team sports analysis has focused on player/ball tracking, activity recognition, localization, player movement forecasting and team formation identification~ \cite{ibrahim2016hierarchical,felsen2017will,wei2014forecasting,lucey2013representing,thomas2017computer,lu2018lightweight,giancola2018soccernet}. For example, activity recognition models for events with well defined group structures have been presented in \cite{ibrahim2016hierarchical}. Attention models have been used to detect key actors \cite{ramanathan_cvpr16} and localize actions (\eg who is the shooter) in basketball videos. Gaze information of players have been used to select proper camera views \cite{arev2014automatic} from first-person videos. 

Hand-crafted features \cite{chen2016learning,felsen2017will}, deep features \cite{tran2015learning} and semantic features \cite{bialkowski2016discovering} have been used to describe the evolution of multi-person sports for various tasks. Most deep features are extracted from the whole image using supervised learning \cite{chen2018camera}. On the other hand, object-level (\eg image patches of players) features are difficult to learn because of the lack of object-level annotations. Our object appearance features are learned from a siamese network \cite{chopra2005learning} without object-level annotations.

\section{USL Broadcast Dataset}
\begin{table*}[t]
 \centering
 \scalebox{0.9} {
  \begin{tabular}{| l |c| c| c | c| c | c | c |}   
    \hline   
    Dataset & Year& Game type & Length (min.) & \# Game & \# Camera & Camera type & Ground truth \\ \midrule
    APIDIS \cite{daniyal2011multi} & 2011& basketball & 15 & 1 & 5 & static & non-prof. \\   
 
    ADS \cite{chen2013computational} & 2013 & field hockey & 70 & 1 & 3 & PTZ & prof. \\ 
    OMB \cite{foote2013one} & 2013 & basketball & $\sim 16$ & 1 & 2 & PTZ & non-prof. \\ 
    VSR \cite{wang2014context} & 2014 & soccer & 9 & 1 & 20 & static & non-prof. \\ 
    CDF \cite{chen2018camera} & 2018 & soccer & 47 & 1 & 3 & PTZ & hybrid \\ \hline
    USL (Ours) & 2018 & soccer & 94 + 108 & 1+12 & 3 & PTZ & prof. \\
    \hline
  \end{tabular}
  }   
  \caption{\textbf{Dataset comparison.} In our dataset, the $108$ minutes data (column four) is sparsely sampled from total $1,080$ minutes data.}
  \vspace{-0.15in}
   \label{table:dataset_cmp}   
\end{table*}

\begin{figure}[t]
	\centering
	\includegraphics[width=0.9\linewidth]{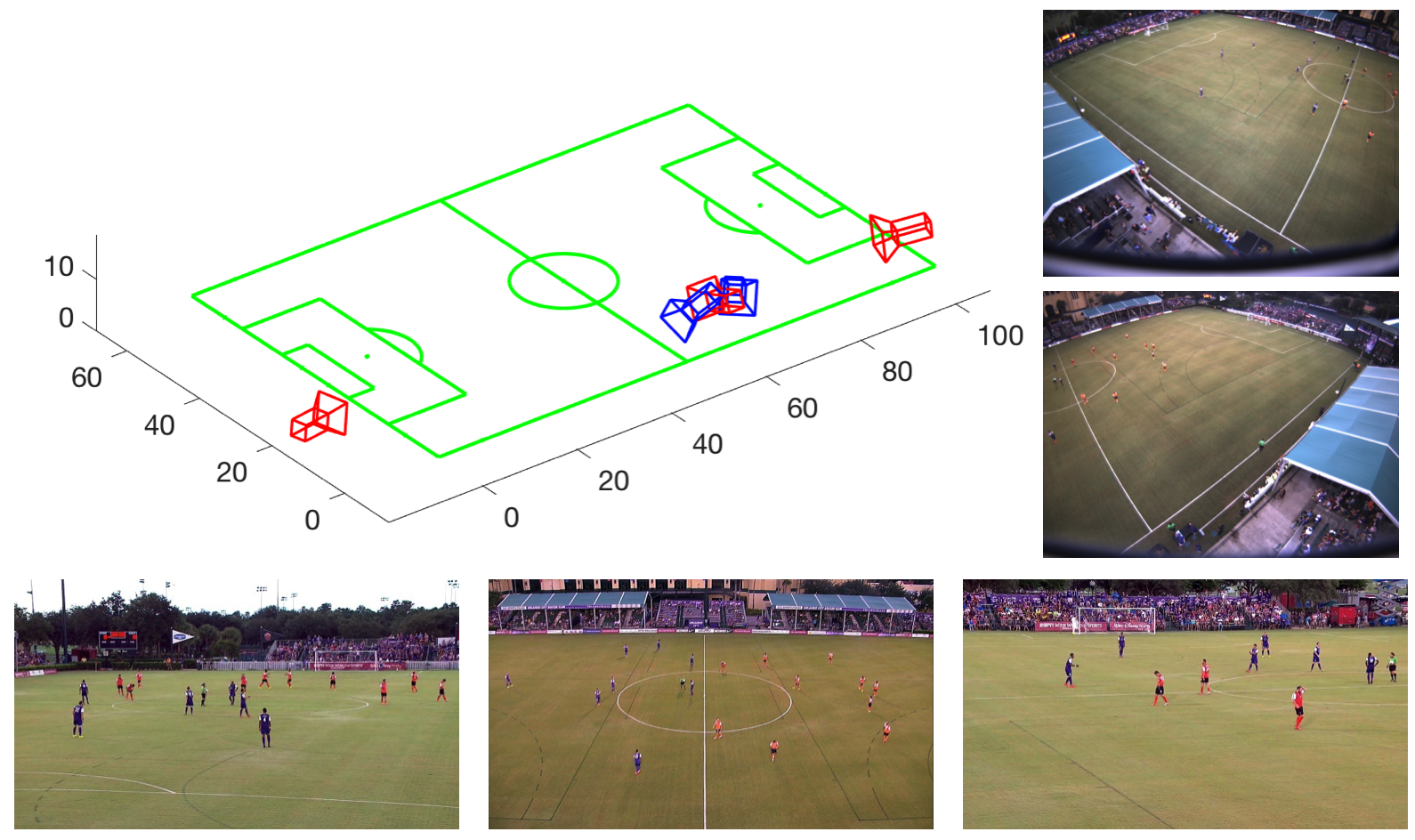}
	\caption{Camera settings of the main game and image examples. Blue: static cameras; red: PTZ cameras. }
	\label{fig:camera_setting}
    \vspace{-0.15in}
\end{figure}

We collect a dataset from United Soccer League (USL) 2014 season. The dataset has one main game and twelve auxiliary games. The main game has six videos. Two videos are from static cameras which look at the left and right part of the playing field, respectively. Three other videos are from pan-tilt-zoom (PTZ) candidate cameras ($1280 \times 720$). Among them, one camera was located at mid-field, giving an overview of the game. The other two cameras are located behind the left and right goals respectively, providing detailed views. Figure \ref{fig:camera_setting} visualizes the camera locations and shows image examples. The sixth video is the broadcast video composited by a professional director from the three PTZ videos. We manually remove commercials and \doubleQuote{replays} and synchronize this video with other videos. The length of the main game is about 94 minutes. Our system only uses information from the three PTZ cameras to select the broadcast camera. Static cameras are only used in one of the baselines. 

Twelve auxiliary games are collected from Youtube. These games are hosted in the same stadium as the main game. They are typically 1.5 hours long. Unlike the main game, each auxiliary game only has a composited broadcast video ($640 \times 360$). Figure \ref{fig:main_idea}(bottom left) shows image examples from the auxiliary games.

In the main game, we manually annotate the ball locations on the static cameras and two detailed view PTZ cameras, at 1fps. In the auxiliary games, we manually check the classified camera IDs around detected camera transition points (details in Section \ref{seq:implementation}). In all games, we detected bounding boxes of players and balls using a method similar to \cite{dai2016r}. Table \ref{table:dataset_cmp} compares our dataset with previous camera view selection datasets. To the best of our knowledge, ours is the first dataset with dense annotations for such long dynamic multi-camera image sequences. We put more dataset details in the supplementary material.

\section{Method}
\label{sec:method}
\begin{figure*}[t]
	\centering
	\includegraphics[width=0.9\linewidth]{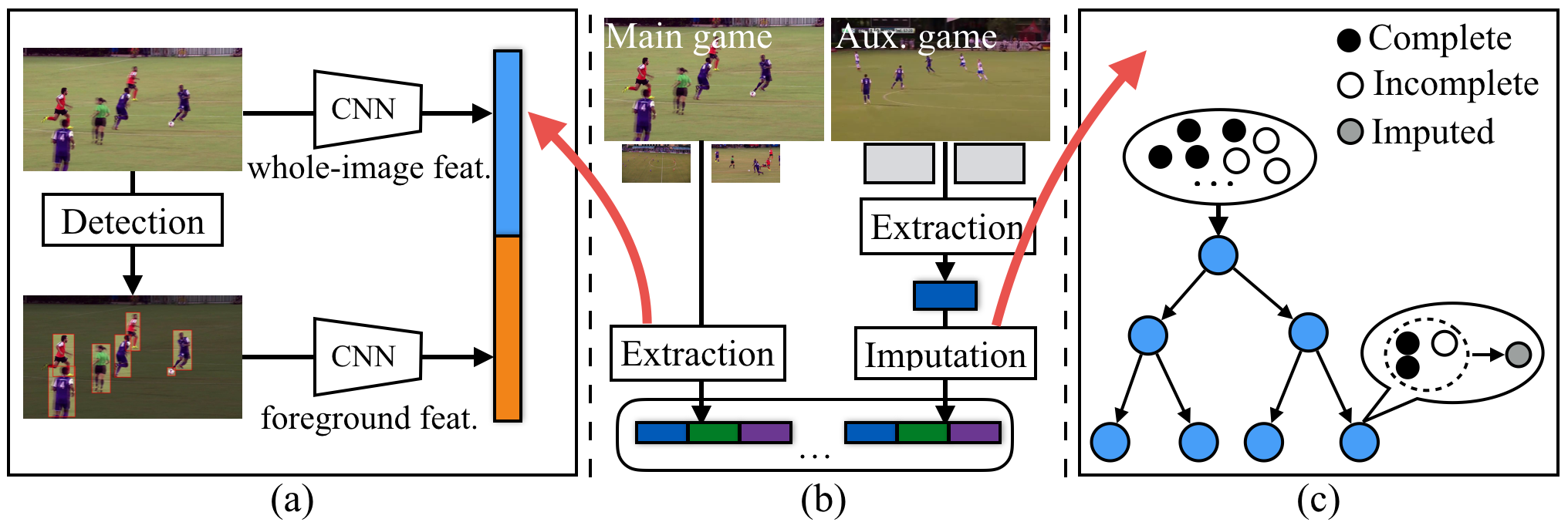}
    \vspace{-0.1in}
	\caption{Main components of our method. (a) Feature extraction. Two CNNs are used to extract whole-image and foreground features. (b) Training process. We first extract features from both main game and auxiliary game frames. The feature of auxiliary games is imputed for the missing data. Both data are then used to train the final model. (c) Data imputation (Section \ref{sec:random_survival_forest}). Best viewed in color. }
	\label{fig:pipeline}
    \vspace{-0.15in}
\end{figure*}
\subsection{Problem Setup}
We have two sources of data. One is from complete games which have videos and selections. Another is from auxiliary games which only have one broadcast video and selections. We model the problem as a classification task given hybrid data $D = \{D_{com}, D_{incom}\}$ in which $D_{com}$ is the \textit{complete} data and $D_{incom}$ is the \textit{incomplete} data. Let $D_{com} = \{X_{com}, Y\}$ where $X_{com}$ is the feature representation of all candidate views and $Y \in \{1,2,3\}$ is the corresponding label. $X$ can be an arbitrary feature representation for an image. Let $D_{incom} = \{\{X_{obs}, X_{mis}\}, Y\}$ where $X_{obs}$ is the \textit{observed} data and $X_{mis}$ is the \textit{missing} data (\eg unrecorded views). Our goal is to learn a classifier from the whole data to predict the best viewpoint from multiple candidate viewpoints (\eg an unseen $X_{com}$):

\begin{equation}
y_t = f(\mathbf{x}_t).
\vspace{-0.05in}
\end{equation}
We do instantaneous single frame prediction and $\mathbf{x}_t$ is a feature representation from all camera views. During training, $\mathbf{x}_t$ is either a raw feature extracted from the main game, or a raw plus imputed feature from an auxiliary game. We only test on the main game. 

Our primary novelty is to use auxiliary data from the Internet which augments the training data with lots of positive examples. On the other hand, this choice creates considerable challenges because of the missing data. 

\vspace{-0.1in}
\paragraph{Assumptions and Interpretation} Our method has three assumptions. First, ${X}_{inputed} = \{X_{obs}, \hat{X}_{mis}\}$ ($\hat{X}$ means the inferred values) and $X_{inputed}$ has a similar distribution as $X_{com}$. This assumption is reasonable since both types of games are collected from the same stadium with the same host team. Also, we expect the broadcast crew to have consistent behaviors across games to some extent. Second, images from different viewpoints are correlated at a particular time instance. Camera operators (from different viewpoints) cooperate to tell the story of the game. For example, often the focus of attention of the cameras is the same (\ie joint attention). In this case, the observed data $X_{obs}$ has a strong indication of the missing data $X_{mis}$. Third, our method models the viewpoint prediction problem as single frame prediction problem without using temporal information. Single-frame prediction is the focus of our work. We will briefly show the adaptation of our method to a temporal model in the experiment.

\subsection{Random Survival Forest}
\label{sec:random_survival_forest}
With these assumptions, we first impute missing data in training. We randomly draw imputed data from the joint posterior distribution of the missing data given the observed data \cite{valdiviezo2015tree}.

\begin{equation}
X_{mis}  \sim p(X_{mis}|X_{obs}, Y)
\vspace{-0.05in}
\end{equation}
with 
\begin{equation}
\label{equ:mis_obs}
p(X_{mis}|X_{obs}, Y) =\int{p(X_{mis}|X_{obs}, \theta)p(\theta|X_{obs}, Y)} \text{d}\theta,
\vspace{-0.05in}
\end{equation}

\noindent where $\theta$ is the model which is decision trees in our method and $Y$ is the label. Please note this process is in training phase so that $Y$ is available.  However, it is often difficult to draw from this predictive distribution due to the requirement of integrating over all $\theta$. Here we introduce random survival forests to simultaneously estimate $\theta$ and draw imputed values.

\begin{figure}[t]
	\centering
	\includegraphics[width=0.9\linewidth]{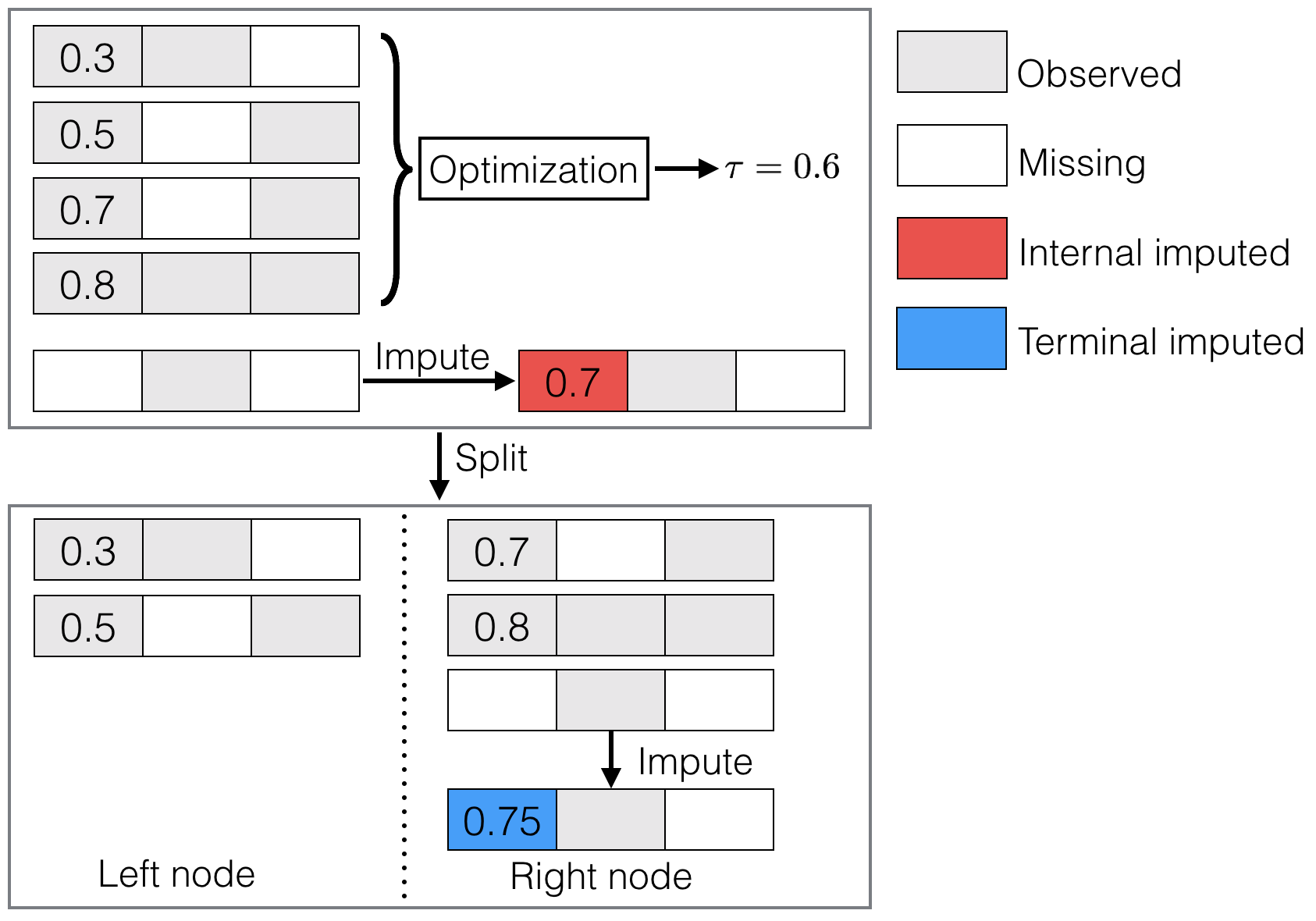}
	\caption{A two-level random survival tree. Each row represents a three-dimensional feature. The first dimension of the fifth feature is imputed. $\tau$ is the decision boundary. Labels are omitted for clarity. Best viewed in color. }
	\label{fig:rst}
    \vspace{-0.15in}
\end{figure}

A random survival forest (RSF) is an ensemble of random survival trees, which was originally designed to identify a complex relationship between long-term survival and attributes of persons (\eg body mass, kidney function and smoking). Each decision tree recursively splits training data into sub-trees until the stopping criteria is satisfied. The statistics (\eg mean values of labels for regression) of training examples in the leaf nodes are used as the prediction \cite{criminisi2012decision}. A survival tree imputes missing data as below.  
\begin{enumerate}
\item In internal nodes, only observed data is used to optimize tree parameters such as the decision boundary by minimizing the cross entropy loss. This step estimates the model $\theta$ from the distribution $p(\theta|X_{obs}, Y)$ in \eqref{equ:mis_obs}.
\item To assign an example with missing data to the left or right sub-trees, the missing value is \doubleQuote{imputed} by drawing a random value from a uniform distribution $U(x|a,b)$ where $(a, b)$ are the lower/upper bounds of $X_{obs}$ of the target dimension. This step draws samples from $p(X_{mis}|X_{obs}, \theta)$ in \eqref{equ:mis_obs}.
\item After the node splitting, imputed data are reset to missing and the process is repeated until terminal nodes are reached. 
\item Missing data in terminal nodes are then imputed using non-missing terminal node data from all the trees. For categorical variables, a majority vote is used; a mean value is used for continuous variables.
\end{enumerate}
Figure \ref{fig:rst} shows the data imputation in a two-level random survival tree. Specific details of RSF can be found in \cite{ishwaran2008random,tang2017random}. With the RSF method, we impute the missing data with substituted values to obtain the new data $\{X_{imputed}, Y\}$. To the best of our knowledge, we are the first to introduce RSF from statistical learning to solve vision problems. Besides, we will experimentally show that it outperforms other alternatives in our problem.

\subsection{Foreground Feature}
\label{subsec:sa_heatmap}
\begin{figure}[t]
	\centering
	\includegraphics[width=0.95\linewidth]{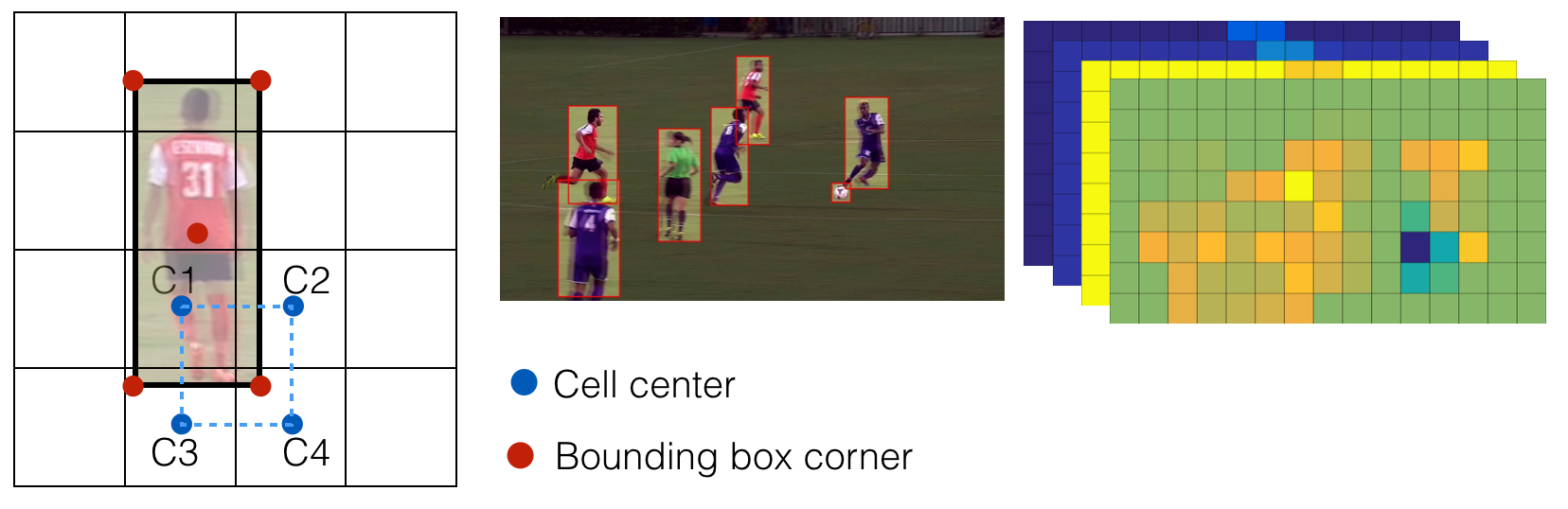}
	\caption{Spatial-appearance heatmap. Left: one player on a $4 \times 4$ grid; right: an example of detected objects and corresponding heatmap.}    
	\label{fig:heatmap}
    \vspace{-0.2in}
\end{figure}
Besides the whole-image feature from a CNN, we also represent foreground objects in an image using a spatial-appearance (SA) heatmap which encodes object appearances in a quantized image space. First, we quantized the image space into a $16 \times 9$ grid. Then, we represent the location of each player using five points (four corners and one center point) of its bounding box. Each point contributes \doubleQuote{heat} to its located and neighboring cells. In the conventional heatmap, the \doubleQuote{heat} is pre-defined values such as the number of players \cite{chen2016learning}. In our heatmap, the \doubleQuote{heat} is the object appearance feature that is learned from the data.

Figure \ref{fig:heatmap} (left) illustrates how the SA heatmap is computed on a $4 \times 4$ grid. The bottom right corner of the bounding box contributes the weighted \doubleQuote{heats} to $C_1, C_2, C_3$ and $C_4$. The weights are the barycentric coordinates of the corner with respect to four cell centers. We use the heatmap as input to train a binary classification CNN and its second-last fully connected layer is used as the foreground feature. 

\vspace{-0.1in}
\paragraph{Appearance Feature Learning}  Given the detected bounding boxes of the objects \cite{dai2016r}, we use a siamese network \cite{chopra2005learning} to learn object appearance features. We train the siamese network using the player tracking information between frames and  extract features from image patches of players in testing. To train the network, we obtain positive (similar) examples from tracked players \cite{lu2013learning} in consecutive frames (\eg from frame 1 to frame 2). The underlying assumption is that the tracked players in consecutive frames have similar appearance, pose and team membership. Any player not part of a track is likely to be dissimilar. The siamese network minimizes the contrastive loss \cite{hadsell2006dim}:
\begin{equation}
\label{equ:con_loss}
\begin{array}{l}
{L_c}({{\bf{x}}_i},{{\bf{x}}_j},{y_{i,j}}) = \\
{y_{i,j}}D{({{\bf{x}}_i},{{\bf{x}}_j})^2} 
 + (1 - {y_{i,j}})\max {(\delta  - D({{\bf{x}}_i},{{\bf{x}}_j}),0)^2},
\end{array}
\vspace{-0.05in}
\end{equation}

\noindent where $\bf{x}_i$ and $\bf{x}_j$ are sub-images, $y_{i, j}$ are similar/dissimilar labels, $D(.)$ is the $L_2$ norm distance and $\delta$ is a margin (1 in this work). The loss function minimizes the distance between paired examples when $y_{i,j} = 1$, and maximizes the distance according to the margin $\delta$ when $y_{i,j} = 0$.

\section{Implementation}
\label{seq:implementation}
\paragraph{Label Estimation of Internet Videos} We pre-process Internet videos for training labels. Given a raw video, we first detect shot boundaries using \cite{ekin2003automatic}. We call the consecutive frames at the shot boundary \textit{boundary frames} for simplicity. Given boundary frames, we train a CNN to classify their camera IDs to four categories (\ie left, middle, right and other-view). The other-view images are commercials, replay logos or frames that are captured from other viewpoints. To train the camera-ID CNN, we first randomly sample 500 training frames from each PTZ video of the main game. For the other-view, we sample the same number of images from a non-sports video. Then, we apply the trained model to classify boundary frames. The classification result is manually checked and refined. The refined boundary frames are used to re-train the CNN. This process is repeated for each video. After five games, the prediction accuracy is about $85\%$. We found this performance is sufficient to lighten the workload of human annotation. Initialized by the CNN then manually corrected, we collect $1,634$ pairs of boundary frames from twelve videos. 

\vspace{-0.1in}
\paragraph{Feature Extraction} Each frame is represented by two types of features: the whole-image feature and the foreground feature. The whole-image feature (16 dimensions) is from a binary classification CNN to classify if an image is selected or not by human operators. The foreground feature (16 dimensions) is described in Section \ref{subsec:sa_heatmap}. We balance the number of positive and negative examples in training. For the main game, we choose the positive candidate view and one of the negative camera views at sampled times. For the auxiliary games, we randomly sample negative examples from the main game. 

\vspace{-0.1in}
\paragraph{Data Imputation and Final Model Training} In data imputation, we randomly sampled $4,000$ frames around camera shot boundaries (within 2 seconds). The imputed data are verified by a model trained from the complete examples (about $2,100$ data passed verification). We use the random forest method to fuse features from all candidate cameras since it is relatively easy to train. In the final model, about $6,000$ examples are uniformly downsampled (1fps) from the main game. The dimension of the feature is 96 ($16 \times 3 \times 2$ for two types of features from three candidate cameras). The parameters of the random forest are: tree number 20, maximum depth 20 and minimum leaf node number 5. More details of the implementation are provided in the supplementary material.

\section{Evaluation}
We evaluate our method on the the USL dataset. To effectively use the data, we test on the main game using 3-fold leave-one-sequence-out cross-validation. We use the camera selection from human operators as the ground truth and report the classification accuracy. We also report the prediction accuracy on the dataset from \cite{chen2018camera} (the first 4 datasets in Table \ref{table:dataset_cmp} are not publicly available). 

\begin{table*}[t]
\centering
  \begin{tabular}{| l |ccc |c |ccc |c |}   
    \hline   
      & \multicolumn{4}{c|}{Main} & \multicolumn{4}{c|}{Main + aux.}  \\ \cline{2-9} 
    Feature        & L & M & R & All & L & M & R & All   \\ \midrule
    whole-image     & 53.4  & 74.4  & 57.5 & 63.2 & 62.8  & 77.8  & 61.4 & 68.5  \\   
    foreground       & 45.9 & 84.1 & 39.5 & 59.7 & 53.1 &  \textbf{86.2}  & 41.3 & 63.2  \\ \hline
    both            & 58.3 & 78.0 & 58.7 & 66.5 & \textbf{70.0} & 85.2 & \textbf{68.9} & \textbf{75.9} \\         
    \hline
  \end{tabular}
  \vspace{1mm}
  \caption{Selection accuracy using different features and training data. \doubleQuote{Main} and \doubleQuote{Main+aux.} mean the training data is from the main game only and is with auxiliary videos, respectively. L, M and R represent the camera on the left, middle, and right side, respectively. The highest accuracy is highlighted by bold.}
   \label{table:main_result}
\end{table*}

For comparison, we also implement six baselines. \textbf{Baseline 1}: constantly select one camera with the highest prior in the training data. This baseline always selects the overview (middle) camera. \textbf{Baseline 2}: select the camera that is closest to the human-annotated grounded truth location of the ball. \textbf{Baseline 3}: predict the camera using the team occupancy map introduced in \cite{bialkowski2013recognising}. The team occupancy map describes the player distribution on the playing ground using tracked players from the static cameras. \textbf{Automated director assistant (ADA)}~\cite{chen2013computational}: it learns a random forest classifier using player distribution and game flow at a time instance. Our implementation augments temporal information by concatenating the features in a 2-second sliding window, making the predictions more reliable. \textbf{C3D}~\cite{tran2015learning}: it is a deep CNN modified from the original C3D network. First, images from three cameras pass through the original C3D network, separately. Then their $fc6$ activations are concatenated and fed into a fully connected network ($1024 \times 32 \times 3$) with \textit{Softmax} loss. \textbf{RDT+CDF}~\cite{chen2018camera}: it uses the recurrent decision tree (RDT) method and a cumulative distribution function (CDF) regularization to predict camera selections in a sequence. Because \cite{chen2018camera} requires real-valued labels in training, we only compare with it on the dataset from \cite{chen2018camera}.

\begin{figure}[t]
	\centering
	\includegraphics[width=1.0\linewidth]{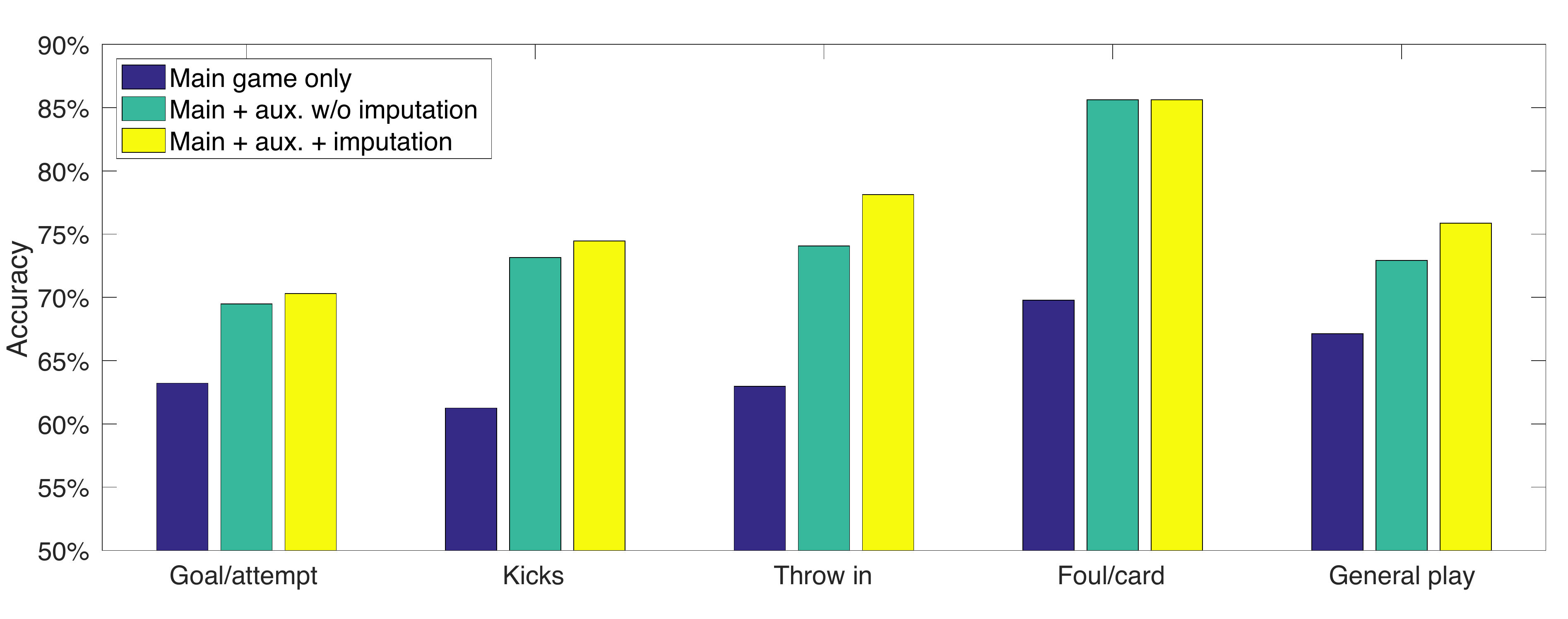}  
    \vspace{-0.2in}
	\caption{Prediction accuracy with and without auxiliary games (grouped by events).}
	\label{fig:accuracy_event}
    \vspace{-0.15in}
\end{figure}
 
\subsection{Main Results}
Table \ref{table:main_result} shows the main results of our method. First, auxiliary data provides significant performance improvement (about 9.4\%). The improvement is from two stages: feature extraction and data imputation. Figure \ref{fig:accuracy_event} shows details of the improvement by separating these two stages and grouping the frames into different events. Overall, the main improvement is from the feature extraction stage (about 6.6\%). Data imputation provides an extra 2.8\% improvement, which is significant in \doubleQuote{throw in} and \doubleQuote{general play}. Second, the foreground feature improves performance, especially when the auxiliary games are used. The main reason might be that the foreground feature excludes the negative impact of backgrounds (\eg different fan groups, weather and light conditions) of auxiliary games.

\begin{table}[t]
 \centering
  \scalebox{1.0}{
  \begin{tabular}{| l | c |c |}   
    \hline
        &  {Accuracy (\%)}  & $\Delta$ \\ \hline
        Constant selection               & 40.9     & 35.0   \\
        Closest to ball (GT.)           & 37.6     &  38.3    \\
    Team occupancy map \cite{bialkowski2013recognising}     & 49.8     &  26.1    \\
    ADA \cite{chen2013computational}     & 54.1 & 21.8 \\
    C3D \cite{tran2015learning}           & 64.3  & 11.6 \\ \hline
    Both feature w/ aux. (Ours)          &  \textbf{75.9}  &  --  \\
    Both feature w/o aux.          & 66.5  & 9.4   \\ 
    whole-image feature w/ aux.     & 68.5  &  7.4  \\   
    foreground feature w/ aux.      & 63.2  &  12.7  \\  
    \hline
  \end{tabular}
  }  
  \caption{Comparison against various baselines and analysis of the effects of various components.}
   \label{table:vs_others}
   \vspace{-0.15in}
\end{table}

\begin{figure}[t]
	\centering
	\includegraphics[width=1.0\linewidth]{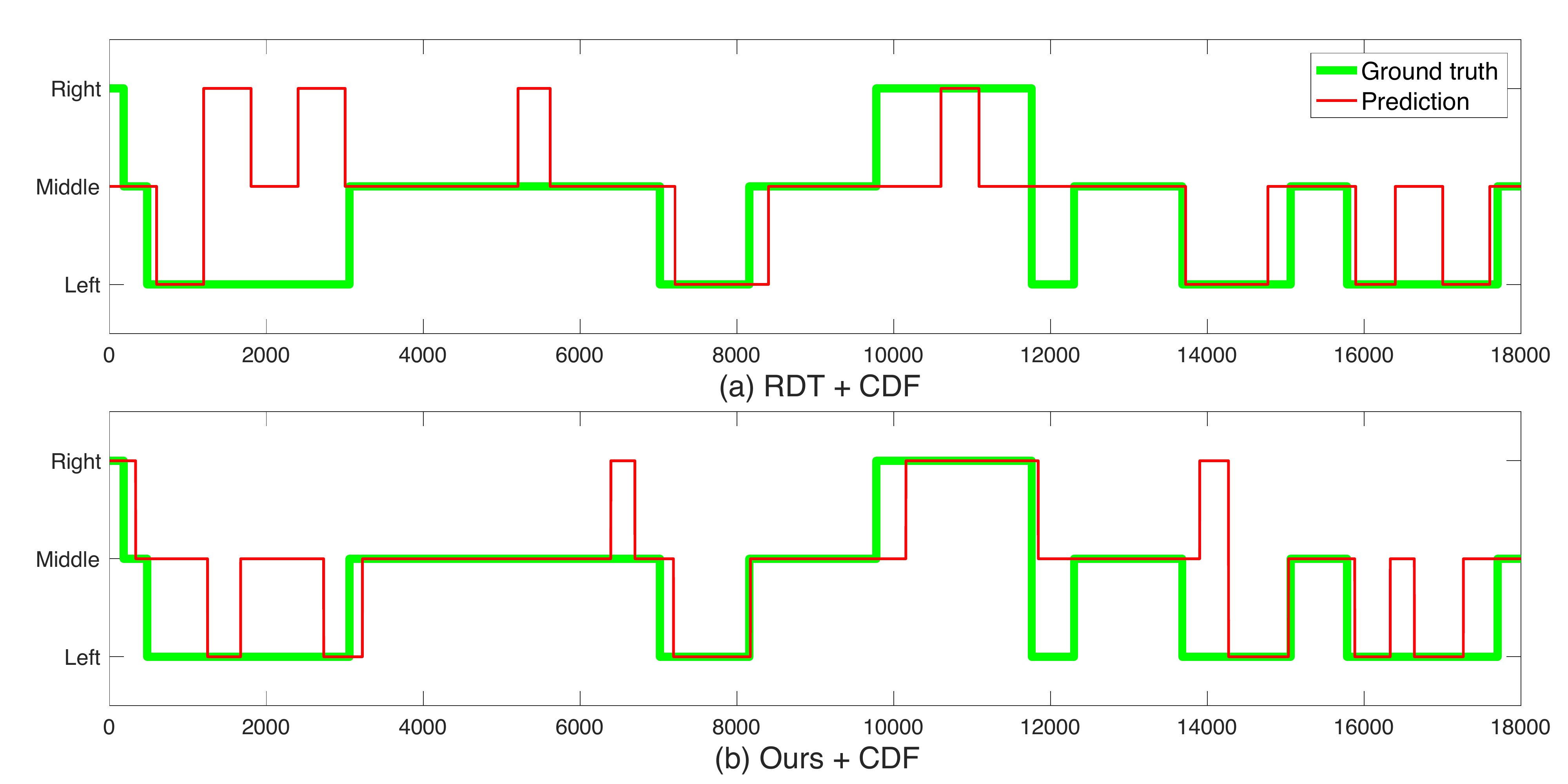}  
    \vspace{-0.2in}
	\caption{Camera selection on a 5-minute testing sequence. (a) Result of \cite{chen2018camera}. (b) Ours. The green line is from human operators and the red line is from algorithms. Best viewed in color.}
	\label{fig:vs_rdt}
    \vspace{-0.15in}
\end{figure}

Table \ref{table:vs_others} shows the comparison with the baselines. Our method outperforms all the methods by large margins. Baselines 1-3 do not work well on this dataset mainly because they omit the image content from the PTZ cameras. This result suggests that heuristic techniques using ball and player locations are not enough for dynamic PTZ camera selection. ADA \cite{chen2013computational} seems to have substantial challenges on this dataset. It is partially because of hand-crafted features (such as player flow) are quite noisy from fast-moving PTZ cameras. C3D \cite{tran2015learning} works reasonably well as it learns both appearance and motion features end-to-end. Its performance is slightly better than our whole-image-feature model. However, our full model is significantly more accurate (11.6 \%) than C3D. It is worth noting that training C3D with auxiliary data is very difficult because the input of C3D is consecutive frames. 

\begin{figure}[t]
	\centering
	\includegraphics[width=1.0\linewidth]{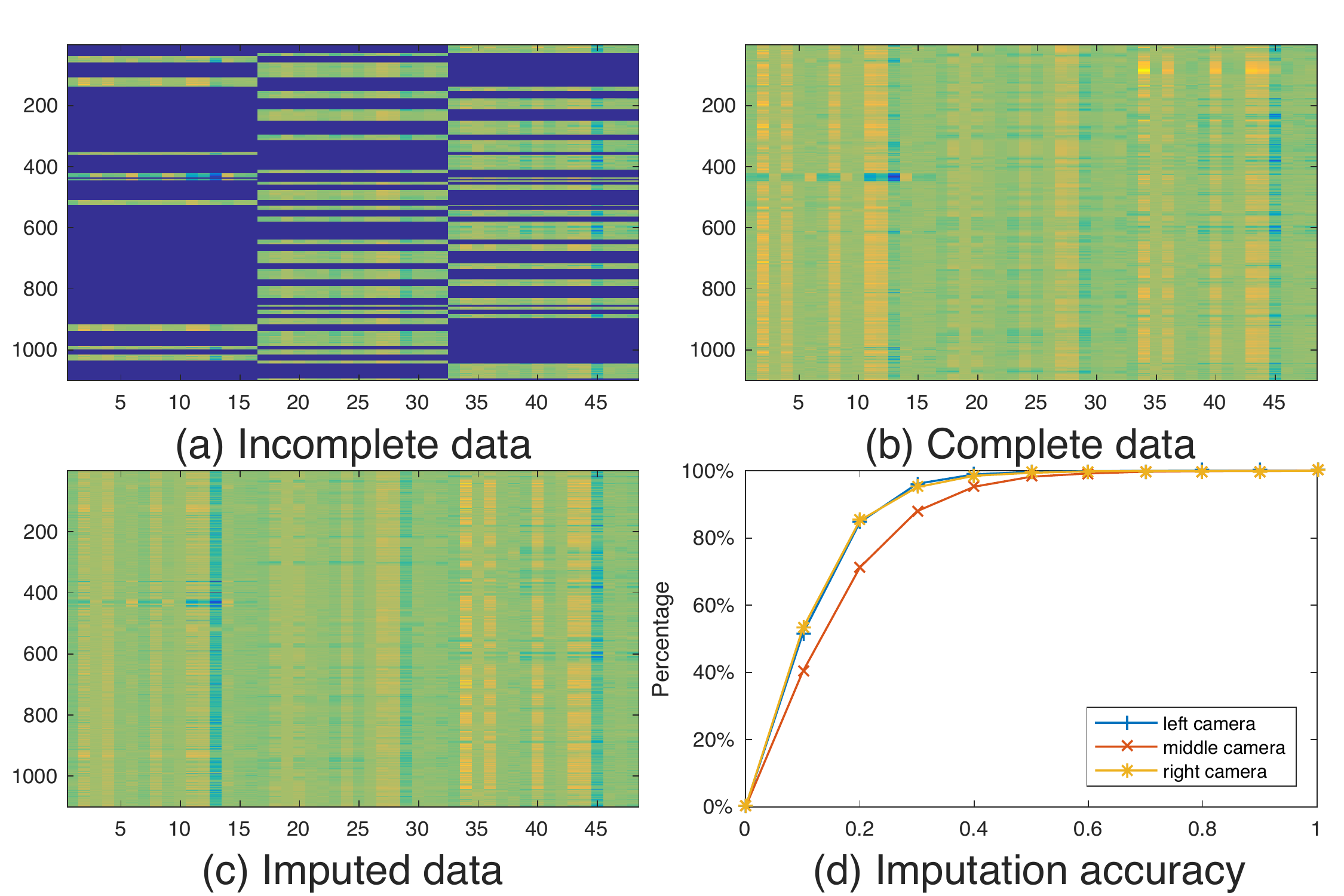}
	\caption{Imputation test on the main game data. (a),(b),(c) Color-coded visualization of imputed feature sequence. The x-axis is the feature dimension. The y-axis is the frame index. Colors visualize the feature values. Blue color blocks indicate the missing data. (d) The imputation accuracy as a function of the error thresholds. Best viewed in color.}
	\label{fig:imputation}
    \vspace{-0.1in}
\end{figure}

\vspace{-0.1in}
\paragraph{Combined with a Temporal Model}
To test the capability of our method with temporal models, we conducted experiments on the dataset from \cite{chen2018camera}. This dataset has 42 minutes (60 fps) data for training and a 5-min sequence for testing. In the experiment, we feed the selection probability to the cumulative distribution function (CDF) method from \cite{chen2018camera}. The CDF method prevents too short (brief glimpse) and too long (monotonous selection) camera selections. The experiment shows our method is more accurate than \cite{chen2018camera} (70\% vs. 66\%). Figure \ref{fig:vs_rdt} shows a visualization of the camera selection. Video results are in the supplementary material for visual inspection.

\subsection{Further Analysis}
\paragraph{Data Imputation Accuracy} Because the missing data in the auxiliary videos has no ground truth, we analyze the accuracy of our data imputation method using the main game data. We use the last $1,100$ frames as testing data by masking the features from the un-selected cameras as missing data. A random survival forest model is trained from the rest of the data. The error is measured by absolute errors normalized by the range of feature in each dimension. This error metric is a good indication of the performance of imputed data in the final model. The final model (\ie a random forest) uses the sign of the difference between the feature value and the decision boundary in internal nodes to guide the prediction. Figure \ref{fig:imputation}(a)(b)(c) visualizes the incomplete data, complete data and imputed data, respectively. The imputed data is visually similar to the ground truth. Figure \ref{fig:imputation}~(d) shows the imputation accuracy as a function of the error thresholds. When the error threshold is 0.2, about $80\%$ of the data are correctly predicted. Although the accuracy is tested on the main game, it suggests a reasonably good prediction on the auxiliary games. 

\begin{table*}[t]
\parbox{.45\linewidth}{
\centering
\begin{tabular}{| l | c |c |}   
    \hline
              & {Acc. (\%)}  & $\Delta$ \\ \midrule        
     RSF           & \textbf{75.9}  &  --  \\  \hline 
     NN                 &   72.2   &  3.7\\
    OptSpace \cite{keshavan2010matrix}   &   68.6   &  7.3 \\
    Autoencoder        &   73.9   &  2.0 \\
    \hline
  \end{tabular}
  \vspace{1mm}
\caption{Comparison of RSF with alternatives.}
\label{table:imputation}
}
\hspace{0.1in}
\parbox{.5\linewidth}{
\centering
\begin{tabular}{| l | c | c|c |c |}   
    \hline
           &  loc.   &  appe.   & {Acc. (\%)}  & $\Delta$ \\ \midrule        
    SA heatmap               & \checkmark & \checkmark &  \textbf{59.7}  &  --  \\ \hline
    Avg pool.          &  & \checkmark & 41.8  &   17.9 \\ 
    Max pool.         &  & \checkmark &  42.4 & 17.3   \\
    Heatmap in [8]  & \checkmark &  &  48.4 &  11.3  \\ % \hline       
    \hline
  \end{tabular}
  \vspace{1mm}
\caption{Comparison of SA heatmap with alternatives.}
\label{table:vs_other_heatmap}
}
\end{table*}

To evaluate the performance of RSF on the real data, we also imputed the missing values using nearest neighbor (NN), OptSpace \cite{keshavan2010matrix} and a neural autoencoder. Table \ref{table:imputation} shows that RSF outperforms all of them with a safe margin. 
\begin{figure*}[t]
	\centering
	\includegraphics[width=0.97\linewidth]{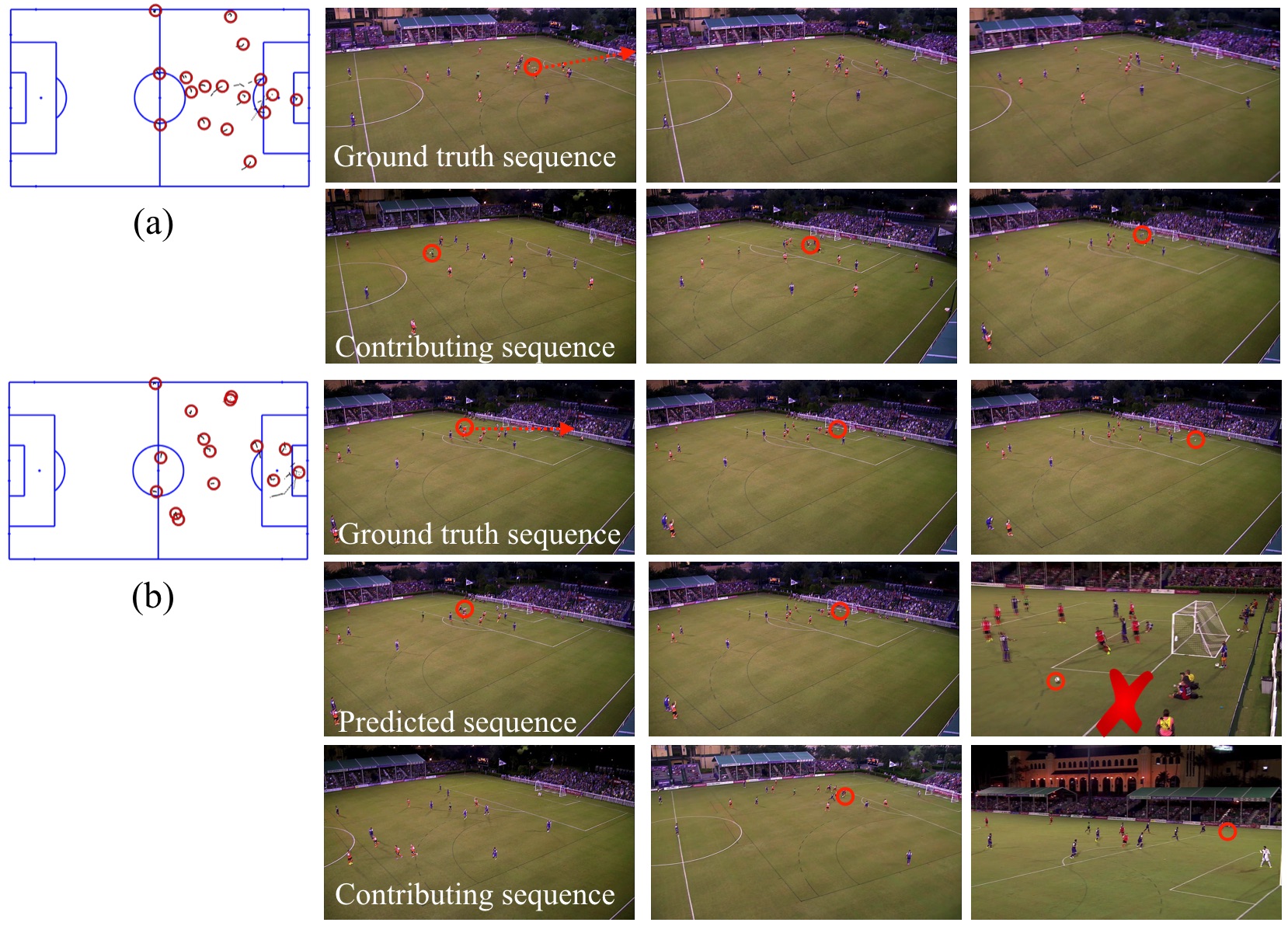}
	\caption{\textbf{Qualitative results.} The ground truth row shows the ground truth image sequences (about 3 seconds). The predicted sequence row shows the predictions from our method (omitted if all predictions are correct such as in (a)). The contributing sequence row shows the most dominant training example in the leaf node. In each sequence, player trajectories are visualized on the playing field template. The ball locations and their trajectories (dashed lines) are overlaid on the original images. The red cross mark indicates incorrect predictions. Best viewed in color.}
	\label{fig:contributing}
    \vspace{-0.1in}
\end{figure*}

\vspace{-0.1in}
\paragraph{Foreground Feature Aggregation} To compare the performance of the SA heatmap with other alternatives, we conducted experiments on the main game data. All the methods use the same appearance feature from the siamese network as input. Table \ref{table:vs_other_heatmap} shows that SA heatmap outperforms other alternatives mainly because it encodes both location and appearance information. 

\vspace{-0.1in}
\paragraph{Qualitative Results}
Figure \ref{fig:contributing} shows predicted image sequences with the ground truth and contributing sequences. The contributing sequence is from the most dominant contributing examples in the leaf nodes for each prediction. Figure \ref{fig:contributing}(b) (last column) shows an example of incorrect predictions. The ground truth camera is kept as the middle camera. Our prediction switches to the right camera. By inspecting the video, we found the human operator's selection has better temporal consistency while ours tends to provide more information in single frames. 

\vspace{-0.1in}
\paragraph{Discussion}
In real applications, more than three candidate cameras are used. However, we found most of the shots are from the three cameras that cover the left goal area, the middle field and the right goal area. We also qualitatively verified that the camera setting in the proposed dataset is representative for soccer games from \cite{wang2014context} and \cite{homayounfar2017sports}. It indicates that our method can be applied to many real situations, especially in small-budget broadcasting. 

Although we collected the largest-ever training data from the Internet, the testing data is from one game. We mitigate this limitation by using dense testing (3-fold cross-validation). We leave large-scale camera selection as future work.

\vspace{-0.1in}
\section{Summary}
In this work, we proposed a framework for sports camera selection using Internet videos to address the data scarcity problem. With effective feature representation and data imputation, our method achieved the state-of-the-art performance on a challenging soccer dataset. Moreover, some of our techniques such as foreground feature extraction are generic and can be applied to other applications. 
The proposed method mainly focuses on camera selection in single frames at the current stage. In the future, we would like to explore temporal information for camera selection.

\noindent {\bf Acknowledgements:} This work was funded partially by the Natural Sciences and Engineering Research Council of Canada. We thank Peter Carr from Disney Research and Tian Qi Chen from University of Toronto for discussions.

{\small
\bibliographystyle{ieee}
\bibliography{egbib}
}

\end{document}